\documentclass[lettersize,journal]{IEEEtran}

\usepackage{cite}
\usepackage{amsmath,amssymb,amsfonts}
\usepackage{graphicx}
\usepackage{textcomp}

\usepackage{multirow}
\usepackage{makecell}
\usepackage{stfloats}
\usepackage{subfig}
\usepackage{epstopdf}
\usepackage{color}
\usepackage{algorithm}
\usepackage{algorithmic}
\usepackage{hyperref}

\usepackage{dsfont}

\begin{document}
		
		\title{CPP-Net: Context-aware Polygon Proposal Network for Nucleus Segmentation}
		
		\author{Shengcong Chen, Changxing Ding, Minfeng Liu, Jun Cheng, and Dacheng Tao,}

		\markboth{ }
		{Chen \MakeLowercase{\textit{et al.}}: CPP-Net: Context-aware Polygon Proposal Network for Nucleus Segmentation}
		
		\maketitle
		
		\begin{abstract}
			Nucleus segmentation is a challenging task due to the crowded distribution and blurry boundaries of nuclei. Recent approaches represent nuclei by means of polygons to differentiate between touching and overlapping nuclei and have accordingly achieved promising performance. Each polygon is represented by a set of centroid-to-boundary distances, which are in turn predicted by features of the centroid pixel for a single nucleus. However, using the centroid pixel alone does not provide sufficient contextual information for robust prediction and thus degrades the segmentation accuracy. To handle this problem, we propose a Context-aware Polygon Proposal Network (CPP-Net) for nucleus segmentation. First, we sample a point set rather than one single pixel within each cell for distance prediction. This strategy substantially enhances contextual information and thereby improves the robustness of the prediction. Second, we propose a Confidence-based Weighting Module, which adaptively fuses the predictions from the sampled point set. Third, we introduce a novel Shape-Aware Perceptual (SAP) loss that constrains the shape of the predicted polygons. Here, the SAP loss is based on an additional network that is pre-trained by means of mapping the centroid probability map and the pixel-to-boundary distance maps to a different nucleus representation. Extensive experiments justify the effectiveness of each component in the proposed CPP-Net. Finally, CPP-Net is found to achieve state-of-the-art performance on three publicly available databases, namely DSB2018, BBBC06, and PanNuke. Code of this paper is available at \url{https://github.com/csccsccsccsc/cpp-net}.
		\end{abstract}
		
		
		\section{Introduction}
		\label{sec:introduction}
		
		\begin{figure}
			\centering
			\includegraphics[width=\linewidth]{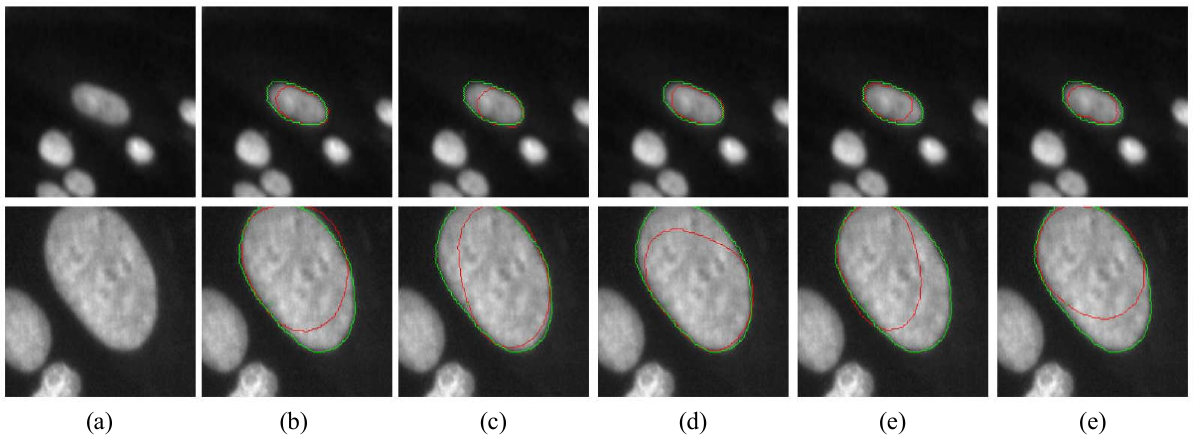}
			\caption{Figures in the first column are original images. The green and red curves in the other five columns denote ground-truth and predicted contours, respectively. Column (b) presents contours predicted by the centroid pixel of one instance \cite{stardist}. The last four columns show contour predictions by pixels in the right-, bottom-, left-, and top-sides of the instance. It is clear that pixels in different locations have complementary pixel-to-boundary distance prediction power.}
			\label{fig:samples_a}
		\end{figure}

		\IEEEPARstart{N}{ucleus} segmentation is a process aimed at detecting and delineating each nucleus in microscopy images. This process is capable of providing rich spatial and morphological information about nuclei; therefore, it plays an important role in many cell analysis applications, such as cell-counting, cell-tracking, phenotype classification and treatment planning \cite{dsb2018}. The segmentation quality affects the measurement of the nucleus shape; therefore, it is essential for applications that depend on the nucleus phenotype \cite{dsb2018, highcontentscreening}. Manual nucleus segmentation is time-consuming, meaning that automatic nucleus segmentation methods have become increasingly necessary.
		
		However, automatic nucleus segmentation still remains a challenging task in terms of robustness due to the crowded distribution of nuclei and their blurry boundaries, as discussed in \cite{stardist}. Unlike objects in natural images \cite{mask-rcnn, mask-ssd, rdcpn}, nuclei tend to overlap with each other. As a result, the bounding box for one instance often covers other nuclei, which negatively impacts the robustness of traditional bounding box-based detection methods, such as Mask R-CNN \cite{mask-rcnn}. Another major challenge lies in the blurry boundary between touching nuclei, which increases the difficulty of inferring their boundaries.
		
		A large number of approaches have been proposed to handle the above challenges \cite{kumar, dcan, hred-net, bes-net, cia-net, tri-unet, rota-net, dist, micro-net, hovernet, mutex_watershed, spa-net, patchperpix, instanceembedding, cnn-dist-tran, brp-net, multistar, maskrcnn_unet, stardist, keygraph, pffn, bpl}. For example, Chen et al. \cite{dcan} differentiate instances of nuclei according to their boundaries. Graham et al. \cite{hovernet} represent nucleus instances using pixel-to-centroid distance maps in both the horizontal and vertical directions. Koohbanani et al. \cite{spa-net} infer nucleus instances by clustering bounding boxes predicted on each pixel within nuclei. When attempting to finally obtain nucleus instances, the above approaches typically resort to complex post-processing operations, such as morphological operations \cite{dcan, cia-net, brp-net}, watershed algorithms \cite{dist,hovernet,mutex_watershed}, and clustering \cite{spa-net}. Several recent works \cite{stardist,polarmask,multistar} represent each instance using a polygon, which is realized by predicting a set of centroid-to-boundary distances. They require only light-weight post-processing operations, i.e., non-maximum suppression, to remove redundant proposals; therefore, their pipelines are more straightforward and efficient.
		
		However, these polygon-based approaches predict polygons using features of the centroid pixel for each instance only, whereas the centroid alone lacks contextual information~\cite{pointAnchor,CNN-GCN}. In particular, the centroid is located far away from boundary pixels for large-sized nuclei, which degrades the distance prediction accuracy. Moreover, supervision is imposed on each respective distance value and there is a lack of global constraint on the shape of each nucleus.
		
		In this paper, we propose a Context-aware Polygon Proposal Network (CPP-Net) to improve the robustness of polygon-based methods \cite{stardist} for nucleus segmentation. The contributions of this paper are made from three perspectives. First, CPP-Net explores more contextual information to improve the prediction accuracy for the centroid-to-boundary distances. As illustrated in Fig. 1, pixels in different locations have complementary pixel-to-boundary distance prediction power. This implies us to promote the accuracy of existing polygon-based methods via exploring the prediction of more pixels inside each instance. Therefore, it adopts the StarDist \cite{stardist} model to conduct initial distance prediction along a set of pre-defined directions. It then samples a set of points between the centroid and the initially predicted boundary along each direction. As these points are closer to the boundary than the centroid pixel, their distance to the ground-truth boundary can be predicted much more accurately. Correspondingly, the initially predicted centroid-to-boundary distance value can be refined with reference to the predictions for those sampled points.
		
		Second, the prediction confidence of these sampled points typically varies according to their feature quality. For example, the errors contained in the distances initially predicted by StarDist \cite{stardist} can be amplified in case where some sampled points actually fall outside the nucleus. Accordingly, the weights of the sampled points should change depending on their prediction confidence. We therefore propose a Confidence-based Weighting Module (CWM) that adaptively fuses the predicted distances for these points. With the assistance of CWM, CPP-Net can more robustly utilize contextual information from the sampled points.
		
		Third, we introduce a novel Shape-Aware Perceptual (SAP) loss, which constrains CPP-Net's predictions regarding the nucleus shape. The original perceptual loss \cite{perceptual-loss} penalizes the differences in the hidden feature maps of a pre-trained classification network between two input images. To encode the shape information of the nucleus into the perceptual loss, we train an encoder-decoder model that maps the representation of nucleus shape in CPP-Net, i.e., the pixel-to-boundary distance maps and the centroid probability map, to other shape representations, such as nucleus bounding boxes. By being trained in this way, this model is capable of extracting rich shape information related to nuclei. We then adopt the encoder part to extract feature maps for the predictions and the ground-truth output of CPP-Net, respectively. The SAP loss penalizes the differences between these extracted feature maps. In this way, the shapes of nuclei during training are constrained.
		
		The contributions of this paper are briefly summarized as bellow:
		\begin{itemize}
			\item We propose a Context Enhancement Module to contrapuntally sample a point set for each nucleus instance to improve the accuracy of predicted distances and a Confidence-based Weighting Module to adaptively merge the sampled information.
			\item We develop a Shape-Aware Perceptual loss to facilitate model optimization through constraining the predictions regarding the nucleus shape.
			\item We develop a Fine-grained Post-Processing pipeline to further correct the false-positive and false-negative predictions.
			\item We conduct experiments on the DSB2018 \cite{dsb2018}, BBBC006v1 \cite{bbbc006} and PanNuke \cite{pannuke,pannuke_extend} databases, and the experimental results justify the effectiveness of CPP-Net.
		\end{itemize}
		
		The remainder of this paper is organized as follows. Related works on nucleus segmentation are reviewed briefly in Section \ref{sec:related_works}. The proposed methods are described in Section \ref{sec:methods}, while implementation details are presented in Section \ref{sec:experiments}. Experimental results are presented in Section \ref{sec:results}, along with their analysis. Finally, we conclude this paper in Section \ref{sec:conclusion}.
		
		\section{Related Works}
		\label{sec:related_works}
		A number of effective approaches for nucleus segmentation have been proposed. In this section, we divide the recent researches into two categories, namely traditional methods and deep-learning based methods.
		
		Many traditional methods are based on the watershed algorithm \cite{wts, wts_track, wts_mp, cellsegreview, hta_mig}. For example, Malpica et al. \cite{wts} proposed a morphological watershed-based algorithm, which is assisted by means of empirically designed image processing operations. This approach utilizes both intensity and morphology information for nucleus segmentation. However, this is likely to cause over-segmentation, and also results in limitations in the processing of overlapping nuclei \cite{wts_track, wts_mp}. Yang et al. \cite{wts_track} proposed a new marker extraction method based on condition erosion to alleviate the over-segmentation problem. Tareef et al. \cite{wts_mp} proposed a Multi-Pass Fast Watershed method that adaptively and efficiently segments overlapping cervical cells. Moreover, the active contour model (ACM) has also been widely adopted for nucleus segmentation \cite{ac,ac_mps}. For example, Molna et al. \cite{ac_mps} proposed to promote the performance of ACM by exploring prior knowledge, specifically the understanding that nuclei usually have ellipse-shaped boundaries. Other traditional methods, such as level-set \cite{lsf}, template-matching \cite{tm}, and cascade sparse regression chain model \cite{csp}, have also been adopted for nucleus segmentation. For these existing methods, morphology information is usually helpful \cite{icip18, cellculturemodel, ellipticalshape}. For example, touching or overlapping nuclei can be partitioned by looking for points with the maximum curvature values \cite{cellculturemodel} or fitting a Gaussian Mixture Model based on the prior knowledge of the elliptic shape \cite{ellipticalshape}. The common downside of traditional methods is that they typically require hand-crafted features, which depend on human expertise and have limitations in terms of their representation power.
		
		In recent years, deep-learning based approaches have achieved notable success on nucleus segmentation tasks~\cite{kumar, dcan, hred-net, bes-net, cia-net, tri-unet, rota-net, dist, micro-net, hovernet, mutex_watershed, spa-net, patchperpix, instanceembedding, cnn-dist-tran, brp-net, multistar, maskrcnn_unet, stardist, keygraph, pffn, bpl}. These works can be further categorized into two-stage and one-stage methods.
		
		Two-stage methods consist of a detection stage, which locates nucleus instances, and a segmentation stage, which predicts a foreground mask for each instance. One representative method of this kind is Mask R-CNN \cite{mask-rcnn} and its variants \cite{pffn, maskrcnn_unet}, which detect nucleus instances using bounding boxes. However, the shape of nuclei tends to be elliptical, and severe occlusion typically exists between instances; this means that each bounding box may contain pixels representing two or more instances, indicating that bounding boxes may be ultimately sub-optimal for nucleus segmentation \cite{stardist,brp-net}. To handle this problem, SpaNet \cite{spa-net} detects instance centroids and performs semantic segmentation in its first stage. In its second stage, it predicts the bounding box of the associated instance according to the feature of each foreground pixel. Finally, it separates overlapping nuclei by clustering the above pixel-wise predictions using the centroids as clustering centers. Moreover, BRP-Net \cite{brp-net} is also a two-stage network. It includes a detection stage, which generates region proposals based on instance boundaries, and a refinement stage, which refines the foreground area of each instance. Notably, neither SpaNet \cite{spa-net} nor BRP-Net \cite{brp-net} is designed in an end-to-end manner, which increases the complexity of the entire system.
		
		By contrast, one-stage methods adopt a single network. Based on the network prediction, they utilize post-processing operations to obtain nucleus instances. Depending on the network prediction property being utilized, one-stage methods can be further subdivided into classification-based models and regression-based models.
		
		As the name suggests, classification-based models output classification probability maps. Existing works in this sub-category include boundary-based \cite{kumar, tri-unet, rota-net, dcan, hred-net, bes-net, cia-net} and connectivity-based \cite{patchperpix} methods. Boundary-based methods typically include a boundary detection branch and a semantic segmentation branch \cite{dcan, tri-unet, bes-net, cia-net}; for example, DCAN \cite{dcan} constructs two separate decoders for boundary detection and semantic segmentation, respectively. Because these two tasks are related, BES-Net \cite{bes-net} and CIA-Net \cite{cia-net} respectively introduce uni- and bi-directional connections between the two branches. These methods process images in the RGB color space. In comparison, Zhao et al. \cite{tri-unet} leveraged the optical characteristics of Haemotoxylin and Eosin (H\&E) staining, and proposed a Hematoxylin-aware Triplet U-Net, which makes predictions with reference to the extracted Hematoxylin component in the image. By subtracting instance boundaries from the segmentation maps, overlapped nuclei can be separated; the downside of this is that such a subtraction operation may result in a loss of segmentation accuracy \cite{brp-net}. Moreover, we term PatchPerPix \cite{patchperpix} as a connectivity-based method, since the prediction it makes indicates whether a pixel is located in the same instance as each of its neighbors. Due to the advantages it offers in the context of describing the local shape of instances in small patches, PatchPerPix is capable of segmenting instances with sophisticated shapes.
		
		In comparison, the regression-based models output regression maps, e.g., distances or coordinate offsets for each pixel of the input image. For example, HoVer-Net \cite{hovernet} predicts the distances from each foreground pixel to its corresponding nucleus centroid in both the horizontal and vertical directions. It then employs the marker-controlled watershed algorithm as post-processing to obtain nucleus instances. The performance of these approaches is affected by the empirically designed post-processing strategies.
		Recently, Schmidt et al. \cite{stardist} proposed the StarDist approach, which predicts both the centroid probability maps and distances from each foreground pixel to its associated instance boundary along a set of pre-defined directions. In the post-processing step, StarDist generates polygon proposals based on the set of predicted distances for each centroid pixel. Each polygon represents one nucleus instance. In this method, polygons are predicted using the features of the centroid pixel only; as a result, contextual information for large-sized nucleus instances is lacking, which affects the prediction accuracy.
		
		Our proposed CPP-Net is a one-stage method that relates closely to StarDist \cite{stardist}. Compared with existing works \cite{mask-rcnn, hovernet, stardist}, CPP-Net fully takes advantages of contextual information for accurate instance segmentation. Specifically, it improves the robustness of StarDist by integrating rich contextual information from a sampled point set for each centroid pixel. Besides, CPP-Net also adopts a novel Shape-Aware Perceptual loss, that constrains CPP-Net's predictions according to the shape prior of nuclei.
		
		\section{Methods}
		\label{sec:methods}
		
		\subsection{Overview}
		
		\begin{figure*}[!htp]
			\centering
			\includegraphics[width=1\linewidth]{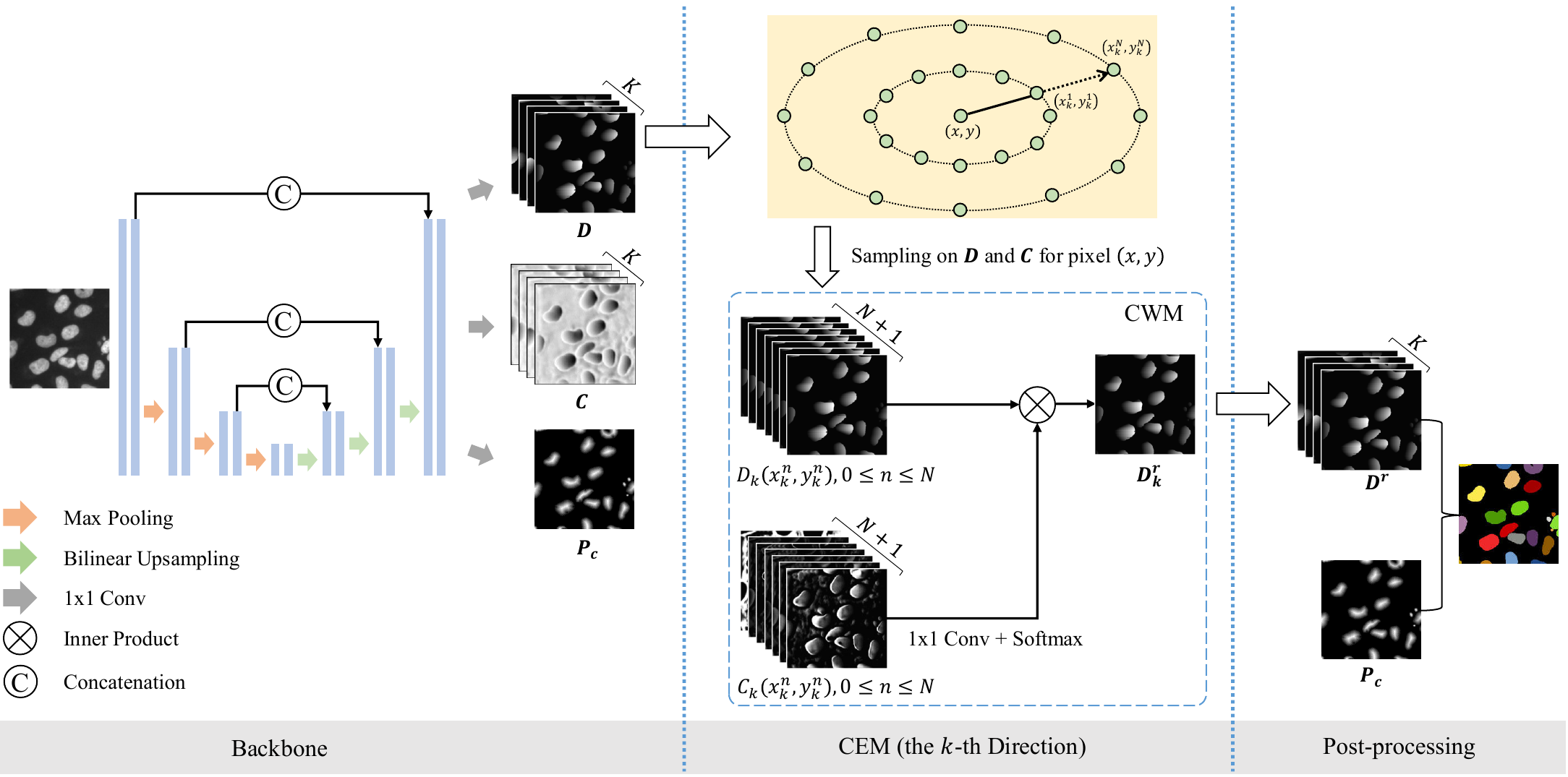}
			\caption{The architecture of CPP-Net. This model adopts U-Net as its backbone, which makes three types of predictions for each input image: the pixel-to-boundary distance maps $\boldsymbol{D}$, the prediction confidence maps $\boldsymbol{C}$, and the centroid probability maps $\boldsymbol{P_c}$. In this figure, we take the $k$-th direction as an illustrative example. The Context Enhancement Module (CEM) conducts sampling on $\boldsymbol{D}$ according to Eq. \eqref{eq:3.1}. Coordinates of the sampled points are computed according to Eq. \eqref{eq:3.2} and Eq. \eqref{eq:3.3}. The Confidence-based Weighting Module (CWM) performs sampling on $\boldsymbol{C}$ in the same location as above. It then produces weights that are used to fuse the distance predictions of the sampled points. In this way, CPP-Net predicts the refined pixel-to-boundary distance maps, i.e., $\boldsymbol{D^r}$, more robustly through the use of rich contextual information. Best viewed in color.}
			\label{overview}
		\end{figure*}
		
		Fig. \ref{overview} presents the structure of CPP-Net for nucleus segmentation. The backbone of CPP-Net is a simple U-Net. Three parallel $1\times1$ convolutional (Conv) layers are attached to the backbone. These layers predict the pixel-to-boundary distance maps $\boldsymbol{D} \in R^{H \times W\times K}$, the confidence maps $\boldsymbol{C} \in R^{H \times W\times K}$, and the centroid probability map $\boldsymbol{P_c} \in R^{H \times W}$, respectively. $H$ and $W$ represent the height and width of the image, respectively. For clarity, we denote the coordinate space of the input image as $\varOmega$ and the total number of elements in $\varOmega$ as $|\varOmega|$. The same as \cite{stardist}, each element in the $k$-th channel of $\boldsymbol{D}$ refers to the distance between a foreground pixel and the boundary of its associated instance along the $k$-th pre-defined direction.
		$K$ denotes the number of total directions.
		Elements in $\boldsymbol{P_c}$ indicate the probability of each foreground pixel being the instance centroid.
		
		In what follows, we first propose a Context Enhancement Module (CEM), which samples a point set to explore more contextual information for pixel-to-boundary distance prediction. We then design a Confidence-based Weighting Module (CWM) that adaptively combines the predictions from the sampled points. Finally, we introduce the Shape-Aware Perceptual (SAP) loss and the Fine-grained Post-Processing (FPP) pipeline, which further promote the segmentation accuracy.

		\subsection{Context Enhancement Module}
		\label{sec:cppnet}
		
		The nucleus segmentation task comprises two subtasks: instance detection and instance-wise segmentation. The recently developed StarDist approach \cite{stardist} performs these two subtasks in parallel. The first detects the centroid of each nucleus, whereas the second segments each instance using a polygon, which is represented using the distances from the centroid pixel to the instance boundary along $K$ pre-defined directions. In \cite{stardist}, the distances are predicted using only the features of the centroid. However, the size of nuclei may vary dramatically, meaning that the centroid pixel alone may lack contextual information for precise distance predictions. 
		
		To handle the above problem, we propose CEM, which utilizes pixels that are closer to the boundaries to refine the distance prediction. To achieve this goal, CEM first samples $N$ points between each pixel and its predicted boundary position along each direction. It then merges the predicted pixel-to-boundary distances of the $N$ points, and adaptively updates the pixel-to-boundary distance of the initial pixel. Formally speaking, the refined pixel-to-boundary distance along the $k$-th direction for one pixel $(x, y)$ can be obtained as follows:
		\begin{equation}\label{eq:3.1}
			D^r_k(x,y)=\sum_{n \in \left[ 0,N \right]}W_k^n(x,y)(D_k(x_k^n, y_k^n)+\frac{n}{N}D_k(x,y)),
		\end{equation}
		where $D_k(x, y)$ denotes the initially predicted pixel-to-boundary distance in $\boldsymbol{D}$ along the $k$-th direction for $(x, y)$. $0 \leq k \leq K-1$, where $k$ indexes the sampling directions. $D_k(x_k^0, y_k^0)$ is equal to $D_k(x, y)$. In this paper, we uniformly sample the $N$ points between the initial pixel and its predicted boundary along each specified direction. The coordinates $(x_k^n,y_k^n)$ for the $n$-th sampled point are accordingly computed as follows:
		\begin{equation}\label{eq:3.2}
			x_k^n=x+\frac{n}{N}D_k(x,y)cos(\frac{2k}{K}\pi),
		\end{equation}
		\begin{equation}\label{eq:3.3}
			y_k^n=y+\frac{n}{N}D_k(x,y)sin(\frac{2k}{K}\pi).
		\end{equation}
		
		Finally, $W_k^n(x,y)$ in Eq. \eqref{eq:3.1} denotes the weight of the $n$-th sampled point. One simple weighting strategy for use is averaging, i.e., setting all $W_k^n(x,y)$ to $\frac{1}{N+1}$. 
		
		\subsection{Confidence-based Weighting Module}
		
		Although the averaging strategy is effective for Eq. \eqref{eq:3.1}, it is also sub-optimal as it neglects the impact of prediction quality on the sampled points. Prediction quality is affected by both image quality and the position of the sampled points. In particular, sampled points near to the boundary may actually lie outside of the nucleus, as $D_k(x,y)$ is only coarse estimation of the pixel-to-boundary distance. Therefore, the prediction accuracy on the sampled points is variable. Accordingly, we propose a Confidence-based Weighting Module (CWM) that adaptively fuses predictions on these sampled points.
		
		As Fig. \ref{overview} illustrates, we attach an extra $1\times1$ convolutional layer to the backbone model in order to produce confidence maps $\boldsymbol{C}$, the sizes of which are the same as those of $\boldsymbol{D}$. Each element in $\boldsymbol{C}$ measures the prediction confidence for the corresponding element in $\boldsymbol{D}$. We then perform sampling on both $\boldsymbol{D}$ and $\boldsymbol{C}$ using coordinates computed according to Eq. \eqref{eq:3.2} and Eq. \eqref{eq:3.3} along each sampling direction, respectively. Sizes of the resulting tensors are therefore $H\times W\times (N+1)$ for each direction. The tensor sampled from $\boldsymbol{C}$ is fed into a $1\times1$ convolutional layer and a Softmax layer.
		The output dimension of the $1\times1$ convolutional layer is also $N+1$. The Softmax layer outputs the normalized weights; these normalized weights are used as $W_k^n(x,y)$ in Eq. \eqref{eq:3.1}. It is worth noting that the $K$ sampling directions share parameters of the $1\times1$ convolutional layer. Therefore, the process of weight generation can be formulated as follows:
			\begin{equation}\label{eq:3.1c1}
				\hat{C}^n_k(x,y) = C_k(x_k, y_k),
			\end{equation}
			\begin{equation}\label{eq:3.1c}
				\boldsymbol{W}_k(x,y)= \sigma(\boldsymbol{\alpha} \cdot \hat{\boldsymbol{C}}_k(x,y) + \boldsymbol{\beta}),
			\end{equation}
			where $\hat{\boldsymbol{C}}_k(x,y)$ denotes $N$ confidences sampled as Eq. \eqref{eq:3.1c1}, while $\boldsymbol{W}_k(x,y)$ denotes the $N$-dimension weight vector on pixel $(x,y)$. $\boldsymbol{\alpha}$ and $\boldsymbol{\beta}$ are the weights and bias in the $1\times1$ convolutional layer, and $\sigma$ denotes the Softmax layer.
		
		\begin{algorithm}[htb]
			\caption{Sampling}
			\label{alg:sampling}
			\begin{algorithmic}[1]
				\REQUIRE ~~\\
				Coordinates for sampling $\boldsymbol{X} \in R^{H \times W}$ and $\boldsymbol{Y} \in R^{H \times W}$\\
				Candidate features $\boldsymbol{F} \in R^{H \times W}$\\
				\ENSURE ~~\\
				Outputs $\boldsymbol{O} \in R^{H \times W}$
				\\[3pt]
				\FOR {$(x,y) \in \varOmega $}
				\STATE $O(x,y)$ = $F(X(x,y), Y(x,y))$
				\ENDFOR
				\RETURN $\boldsymbol{O}$;
			\end{algorithmic}
		\end{algorithm}
		
		\begin{algorithm}[htb]
			\caption{Context Enhancement Module}
			\label{alg:CEM}
			\begin{algorithmic}[1]
				\REQUIRE ~~\\
				The distance predictions $\boldsymbol{D} \in R^{H \times W\times K}$;\\
				The confidence maps $\boldsymbol{C} \in R^{H \times W\times K}$;\\
				The number of sampling points $N$;
				\ENSURE ~~\\
				Refined distance predictions $\boldsymbol{D^r} \in R^{H \times W\times K}$;
				\\[3pt]
				\STATE Calculate the Cartesian coordinates of the $N+1$ sampling points according to Eq. \eqref{eq:3.2} and Eq. \eqref{eq:3.3}, and obtain $\boldsymbol{X} \in R^{H \times W \times N \times K}$ and $\boldsymbol{Y} \in R^{H \times W \times N \times K}$;
				\FOR {$k=1$ to $K$}
				\FOR {$n=1$ to $N$}
				\STATE $\boldsymbol{\hat{D}}_k^n \gets Sampling(\boldsymbol{X}_k^n, \boldsymbol{Y}_k^n, \boldsymbol{D}_k)$;
				\\[2pt]
				\STATE $\boldsymbol{\hat{C}}_k^n \gets Sampling(\boldsymbol{X}_k^n, \boldsymbol{Y}_k^n, \boldsymbol{C}_k)$;
				\ENDFOR
				\ENDFOR
				\STATE Calculate the weights $\boldsymbol{W}$ according to Eq. \eqref{eq:3.1c};
				\STATE Calculate the refined distances $\boldsymbol{D^r}$ according to Eq. \eqref{eq:3.1};
				\RETURN $\boldsymbol{D^r}$;
			\end{algorithmic}
		\end{algorithm}

		\subsection{Loss Functions}
		\label{sec:saploss}
		\begin{figure*}[!htp]
			\centering
			\includegraphics[width=0.99\linewidth]{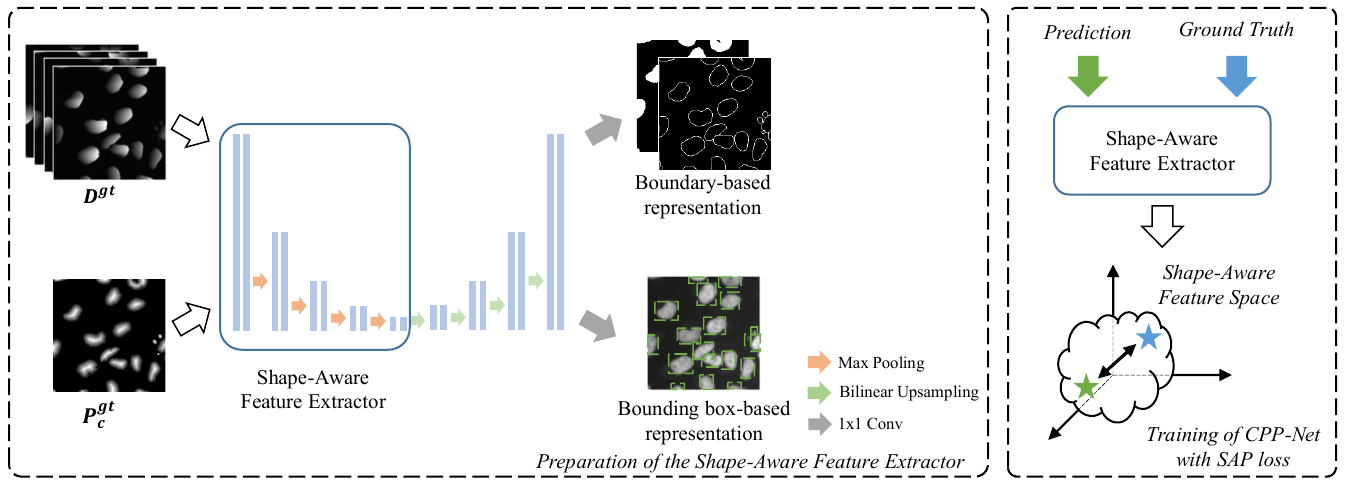}
			\caption{Illustration of the SAP loss. The transformation model in the left sub-figure converts the instance representations utilized in CPP-Net to other forms of instance representation. After the training of the transformation model is completed, the parameters of its encoder are fixed. The encoder can extract high-level shape features of the nuclei; and is therefore used as a shape-aware feature extractor in the SAP loss, as shown in the right sub-figure.}
			\label{fig:iap}
		\end{figure*}
		
		The StarDist model \cite{stardist} utilizes two loss terms: the binary cross entropy loss for centroid probability prediction, and the weighted L1 loss for pixel-to-boundary distance regression. These two loss terms are formulated as follows:
		\begin{equation}\label{eq:3.5a}
			\begin{split}
				L_{prob}=\frac{1}{|\varOmega|}\sum_{(x,y) \in \varOmega} P_c^{gt}(x,y) \log (P_c(x,y))\\
				+(1-P_c^{gt}(x,y))\log(1-P_c(x,y)),
			\end{split}
		\end{equation}
		\begin{equation}\label{eq:3.5b}
			L_{dist}=\frac{1}{K|\varOmega|}\sum_{(x,y) \in \varOmega}{\sum_{k=0}^{K-1}{P_c^{gt}(x,y)| D_k^{gt}(x,y)-D_k(x,y) |}},
		\end{equation}
		\begin{equation}\label{eq:3.5c}
			L_{SD}=L_{prob}+L_{dist},
		\end{equation}
		where $P_c^{gt}(x,y)$ and $P_c(x,y)$ represent elements in the ground-truth and predicted centroid probability maps, respectively. We follow the same process as that outlined in \cite{stardist} to obtain the ground-truth centroid probability map, i.e., utilizing the normalized pixel-to-boundary distance map as centroid probability map. $D_k^{gt}(x,y)$ and $D_k(x,y)$ denote elements of the ground-truth and predicted pixel-to-boundary maps respectively along the $k$-th direction.
		
		For CPP-Net, there are two predicted distance maps, namely $\boldsymbol{D}$ and $\boldsymbol{D^r}$. $\boldsymbol{D}$ is predicted by the backbone model, while $\boldsymbol{D^r}$ represents the final pixel-to-boundary distance prediction by CPP-Net. Accordingly, we modify Eq. \eqref{eq:3.5b} for CPP-Net as follows:
		\begin{equation}\label{eq:3.5d}
			\begin{split}
				L_{dist}'=\frac{1}{K|\varOmega|}\sum_{(x,y) \in \varOmega}\sum_{k=0}^{K-1}P_c^{gt}(x,y)(| D_k^{gt}(x,y)-D_k(x,y) |\\
				+| D_k^{gt}(x,y)-D_k^r(x,y) |),
			\end{split}
		\end{equation}
		where $D_k^r(x,y)$ denotes the refined pixel-to-boundary distance in $\boldsymbol{D^r}$ along the $k$-th direction for $(x,y)$.
		
		Eq. \eqref{eq:3.5b} and Eq. \eqref{eq:3.5d} penalize the prediction error in each respective pixel-to-boundary distance value, while the overall shapes of nucleus instances are ignored. In fact, nucleus instances typically have similar shapes; this can be utilized as the prior knowledge to facilitate accurate nucleus segmentation. However, it is challenging to explicitly represent the overall shape of a single nucleus instance. To deal with this problem, we adopt an implicit approach inspired by the perceptual loss \cite{perceptual-loss}, which is proposed for style transformation and super-resolution tasks. In \cite{perceptual-loss}, a network pre-trained for image classification on ImageNet \cite{imagenet} is used as a feature extractor, with the differences between the extracted features of one image pair being penalized. This approach encourages the high-level information of the two images to be similar. Inspired by the original perceptual loss, we propose a Shape-Aware Perceptual (SAP) loss for nucleus segmentation. In the followings, we introduce the SAP loss in details.
		
		\subsubsection{Preparation of the Shape-Aware Feature Extractor}
		The aim of the SAP loss is to penalize the differences in shape feature between the predicted and ground-truth nucleus representations. To encode the shape information in a deep model, we propose to transform the nucleus representations in CPP-Net, i.e., the pixel-to-boundary distance maps $\boldsymbol{D}$ and the centroid probability map $\boldsymbol{P_c}$, to other representation forms \cite{stardist, cia-net, hovernet, spa-net}. This transformation is accomplished using an encoder-decoder structure as illustrated in Fig. \ref{fig:iap}.
		
		This paper mainly considers two nucleus representation strategies: first, the semantic segmentation and boundary detection maps in boundary-based approaches \cite{cia-net}; second, the location and the size of the associated bounding box for each nucleus. During training of the transformation model, we concatenate the ground-truth $\boldsymbol{D^{gt}}$ and $\boldsymbol{P_c^{gt}}$ for each training image to create the inputs. The binary cross-entropy loss and L1 loss are adopted for the two target representation strategies, respectively.
		
		After the training of the transformation model is completed, we adopt its encoder as the shape-aware feature extractor in the SAP loss. The extractor is denoted as $f_e$ in the following. Parameters of $f_e$ are fixed during the training of CPP-Net.
		
		\subsubsection{Training CPP-Net with the SAP loss}
		
		The SAP is computed as follows:
		
		\begin{equation}\label{eq:3.5g0}
			\boldsymbol{S}=f_e(\boldsymbol{D^{gt}}, \boldsymbol{P_c^{gt}}) -f_e(\boldsymbol{D}, \boldsymbol{P_c}),
		\end{equation}
		\begin{equation}\label{eq:3.5g1}
			\boldsymbol{S^r}=f_e(\boldsymbol{D^{gt}}, \boldsymbol{P_c^{gt}}) -f_e(\boldsymbol{D^r}, \boldsymbol{P_c}),
		\end{equation}
		\begin{equation}\label{eq:3.5g}
			\begin{split}
				L_{SAP}=\frac{1}{|\varOmega'|} \sum_{(x',y') \in \varOmega'}\|\boldsymbol{s}(x',y')\|_1+\|\boldsymbol{s^r}(x',y')\|_1,
			\end{split}
		\end{equation}
		where $\varOmega'$ denotes the 2D coordinate space of the extracted shape-aware feature maps, while $\boldsymbol{s}(x',y')$ and $\boldsymbol{s^r}(x',y')$ are the vectors in $\boldsymbol{S}$ and $\boldsymbol{S^r}$ at the location of $(x',y')$, respectively. The parameters of $f_e$ are fixed during the training of CPP-Net. Finally, the entire loss of CPP-Net is summarized as follows:
		\begin{equation}\label{eq:3.5h}
			L_{CPP}=L_{prob}+L_{dist}'+L_{SAP}.
		\end{equation}
		In the interests of simplicity, we adopt equal weights for the three terms in $L_{CPP}$.

		\subsection{Post-processing}
		\label{fpp}
		As Fig. \ref{overview} illustrates, in the inference stage, $\boldsymbol{D^r}$ and $\boldsymbol{P_c}$ are used to produce each instance mask through post-processing. The post-processing pipeline proposed in StarDist \cite{stardist} comprises two steps: Non-Maximum Suppression (NMS) and conversion from a single polygon to a mask. The NMS step removes redundant polygons obtained from adjacent pixels.
		
\begin{figure}
			\centering
			\includegraphics[width=1.0\linewidth]{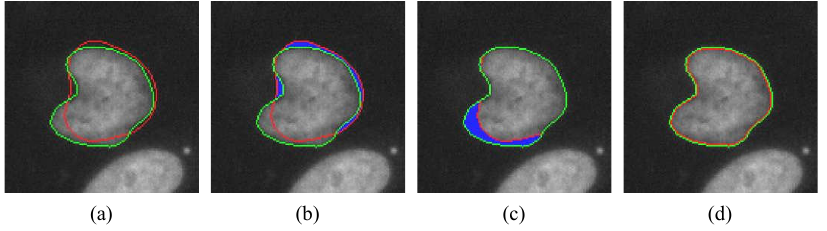}
			\caption{Illustration of FPP. The boundary of the semantic segmentation mask and that of the polygon are colored in green and red, respectively. From (b) to (d), the nucleus boundary colored in red evolves according the semantic segmentation boundary colored in green. The modified pixels in the two steps, i.e., false-positive pixel removal and false-negative pixel assignment, are colored in blue in (b) and (c), respectively. In particular, we assign each false negative pixel to one nucleus instance according to the strategies introduced in Section \ref{fpp}. Best viewed in color.}
			\label{fig:post_processing}
		\end{figure}

		As illustrated in Fig. \ref{fig:post_processing}(a), one polygon only approximates the area for a nucleus instance. To correct the false negative and false positive predictions, we further propose an improved post-processing method named Fine-grained Post-Processing (FPP), which is illustrated in Alg. \ref{alg:FPP} and Fig. \ref{fig:post_processing}(b-d). First, we attach a semantic segmentation decoder to the encoder of CPP-Net. This decoder has the same architecture as the original decoder, and produces a binary mask during inference, which we use to identify all foreground pixels. Second, we execute NMS and convert each obtained polygon to a mask. Third, we remove each background pixel located inside of each polygon, and assign each foreground pixel that lies outside of polygons to one instance. Since the removed pixels are potential false-positive predictions, the former step can further avoid over-segmentation. Moreover, the re-assigned pixels are potential false-negative predictions, the later step can further avoid under-segmentation. In more detail, for each pixel that requires label assignment, we average the coordinates of the $K$ boundary points estimated by the pixel according to $\boldsymbol{D^r}$, which enables us to obtain the centroid coordinate of its associated nucleus instance. Finally, we assign the pixel to one instance according to the estimated centroid coordinates. The modified pixels in the procedures of false-positive pixel removal and false-negative pixel assignment, are colored in blue in Fig. \ref{fig:post_processing}(b) and Fig. \ref{fig:post_processing}(c), respectively.
		
		\begin{algorithm}[htb]
			\caption{Fine-grained Post-Processing}
			\label{alg:FPP}
			\begin{algorithmic}[1]
				\REQUIRE ~~\\
				Refined distance prediction maps $\boldsymbol{D^r} \in R^{H \times W \times K}$;\\
				Semantic segmentation predictions $\boldsymbol{S} \in R^{H \times W}$;\\
				Segmentation threshold value $\tau$
				\ENSURE ~~\\
				Final predictions for instance segmentation $\boldsymbol{L} \in R^{H \times W}$;
				\\[3pt]
				\STATE Perform the post-processing pipeline in StarDist \cite{stardist}, and obtain the initial predictions for instance segmentation $\boldsymbol{L}$;
				\STATE $\boldsymbol{S}^{fg} \gets \mathds{1}(\boldsymbol{S} \ge \tau)$
				\STATE $\boldsymbol{L} \gets \boldsymbol{L} \odot \boldsymbol{S}^{fg}$;
				\STATE Convert $\boldsymbol{D^r}$ to Cartesian coordinates $\boldsymbol{X} \in R^{H \times W \times K}$ and $\boldsymbol{Y} \in R^{H \times W \times K}$;
				\STATE Average $\boldsymbol{X}$ and $\boldsymbol{Y}$ along the direction dimension respectively, and obtain the estimated centroid coordinate maps $\boldsymbol{X^c} \in R^{H \times W}$ and $\boldsymbol{Y^c} \in R^{H \times W}$;
				\STATE $\boldsymbol{L_0} \gets \boldsymbol{L}$
				\STATE $t \gets 0$
				\REPEAT
				\STATE $\boldsymbol{\hat{L}} \gets Sampling(\boldsymbol{X^c}, \boldsymbol{Y^c}, \boldsymbol{L})$;
				\STATE $\boldsymbol{L}_{t+1} \gets \mathds{1}(\boldsymbol{L}_{t}>0)\odot\boldsymbol{L}_{t} + \mathds{1}(\boldsymbol{L}_{t}=0)\odot\boldsymbol{S}^{fg}\odot\boldsymbol{\hat{L}} $
				\STATE $ t \gets t+1$
				\UNTIL $\boldsymbol{L}_{t} = \boldsymbol{L}_{t-1}$
				\STATE $ \boldsymbol{L} \gets \boldsymbol{L}_{t}$
				\RETURN $\boldsymbol{L}$;
			\end{algorithmic}
		\end{algorithm}
		
		\section{Experimental Setup}
		\label{sec:experiments}
		
		To justify the effectiveness of CPP-Net, we conduct extensive experiments on publicly available datasets, i.e., DSB2018 \cite{dsb2018}, BBBC006v1 \cite{bbbc006}, and PanNuke \cite{pannuke, pannuke_extend}.
		
		\subsection{Datasets}
		
		\subsubsection{DSB2018}
		Data Science Bowl 2018 (DSB2018) \cite{dsb2018} is a nucleus detection and segmentation competition, in which a dataset of 670 images and manual annotations are available. Image size in DSB2018 varies from $256\times256$ to $520\times696$ pixels. Multiple staining types, e.g., Hoechst 33342 and DAPI, are also adopted. To facilitate fair comparisons with existing approaches, we follow the evaluation protocol outlined in \cite{stardist}. In this protocol, the training, validation, and testing sets include 380, 67, and 50 images, respectively.
		
		\subsubsection{BBBC006v1}
		BBBC006v1\cite{bbbc006} comprises 768 images of $696\times520$ pixels. It contains U2OS cells stained with Hoechst 33342 markers. In our experiments, we randomly divide the dataset into training, validation, and testing subsets, which contain 462, 153, and 153 images, respectively.
		
		\subsubsection{BBBC039}
		BBBC039\cite{bbbc006} consists of 200 images of size $520\times696$ pixels. These images were captured using fluorescence microscopy with the Hoechst stain. In our experiments, BBBC039 is used only as a testing set for cross-dataset evaluation. To avoid data snooping, we remove three images that are also included in DSB2018; as a result, 197 images are employed during testing.
		
		\subsubsection{PanNuke}
		PanNuke\cite{pannuke,pannuke_extend} is an H\&E stained image set, containing 7,904 $256\times256$ patches from a total of 19 different tissue types. The nuclei are classified into neoplastic, inflammatory, connective/soft tissue, dead, and epithelial cells. We follow the evaluation protocol outlined in \cite{pannuke_extend}, which divides the patches into three folds containing 2,657, 2,524, and 2,723 images, respectively. Three different dataset splits are then made based on these three folds.
		One fold of data is used for training, with the remaining two folds used as validation and testing sets, respectively.
		
		\subsection{Implementation Details}
		
		On DSB2018 and BBBC006v1, we adopt a very similar U-Net backbone as that used in \cite{stardist} for CPP-Net to facilitate fair comparison. This backbone includes three down-sampling blocks in its encoder and three up-sampling blocks in its decoder. The only change is that we replace all Batch Normalization (BN) layers \cite{batchnorm} with Group Normalization (GN) layers \cite{groupnorm}, since we use a small batch size of 1 for training. On PanNuke, we make two changes to this backbone. First, to ensure fair comparison with existing approaches \cite{hovernet}, we replace the encoder of this backbone with ResNet-50 \cite{resnet}, and initialize its weights with those pre-trained on ImageNet \cite{imagenet}. Second, we attach another decoder to classify nucleus types for each input image pixel. Loss functions for this decoder include the sum of the Cross Entropy loss and the Dice loss \cite{dice-loss}.
		
		The architecture of the encoder-decoder model adopted in the SAP loss is very similar to the U-Net backbone in CPP-Net. To enable the SAP loss to extract more high-level information, we make two changes. First, we utilize a deeper structure that includes four down-sampling and four up-sampling blocks. Second, we remove shortcuts that usually passes low-level information from the encoder to decoder layers.
		
		The Adam algorithm \cite{adam} is employed for optimization. The initial learning rate is set to $1\times10^{-4}$, and is reduced by multiplying it by 0.5 if the validation loss no longer reduces. The training process halts if the learning rate is reduced to less than $1\times10^{-7}$. We adopt online data augmentation of random rotation and horizontal flipping during training. As for the encoder-decoder model in the SAP loss, we adopt the same data division protocol and use the same training settings outlined above, except that data augmentation is not employed. In the post-processing, the thresholds on the centroid probability map and semantic segmentation map are set to 0.4 and 0.5, respectively. The IoU threshold for the NMS step is set to 0.5.

		\subsection{Evaluation Metrics}
		
		For DSB2018 and BBBC006v1, we adopt the same evaluation metric as in \cite{dsb2018} and \cite{stardist}. According to the metric, the average precision (AP) with IoU thresholds ranging from 0.5 to 0.9 with a step size of 0.05 are computed. Moreover, we also evaluate the performance of different models with Aggregated Jaccard Index (AJI) \cite{aji} and Panoptic Quality (PQ) \cite{pq}. AJI is based on instance-wise IoU between the ground-truth and the prediction. It has been widely used to measure the performance of nucleus segmentation methods \cite{kumar, dist, cia-net, hovernet, brp-net, tri-unet}. PQ takes both the detection quality and instance-wise segmentation quality into consideration, which has been widely adopted in panoptic segmentation tasks \cite{pq} and was introduced into nucleus segmentation in \cite{hovernet}. For the PanNuke database, we also adopt the PQ presented in \cite{pannuke} as the evaluation metric. We report the PQs of all 19 tissues. Besides, both multi-class PQ (mPQ) and binary PQ (bPQ) are computed for evaluation.
		The mPQ averages the PQ performance on each of the five nucleus categories, while the bPQ directly computes the overall performance on images of all five nucleus categories.
		
		\section{Experimental Results}
		\label{sec:results}
		
		In what follows, we first conduct experiments on the validation sets of two publicly available databases, DSB2018 \cite{stardist} and BBBC006v1 \cite{bbbc006}, to determine the optimal number of sampling points $N$ and demonstrate the effectiveness of the CEM module. We then justify the effectiveness of the CWM module, the SAP loss, and the FPP method on the validation sets of DSB2018 and BBBC006v1. Finally, we compare the performance of CPP-Net with other methods on the testing sets of all three databases.
		
		\subsection{Evaluation of CEM}
		
		In this experiment, we evaluate the optimal number of sampling points in CEM. To facilitate clean comparison, we remove the SAP loss for CPP-Net, and consistently adopt CWM as the weighting strategy in Eq. \eqref{eq:3.1}. We also adopt the original post-processing method in StarDist \cite{stardist}. We further change the number of sampling points, i.e., $N$, from 0 to to 10, and report the experimental results in Table \ref{tab:num_sp}. When $N$ is equal to 0, CPP-Net reduces to the StarDist model \cite{stardist}. As Table \ref{tab:num_sp} shows, the performance of CPP-Net continues to improve as $N$ increases from 0 to 6; however, its performance saturates when $N$ exceeds 6. These results indicate that sufficient contextual information can be captured with 6 sampling points. Therefore, we consistently set $N$ to 6 in the following experiments.
		
		It is clear that a single sampling point alone is able to boost the APs on both databases, especially for APs under high IoU thresholds. Moreover, when $N$ is equal to 6, On DSB2018, CEM improves the mean APs by $3.45\%$, the AJI score by $1.87\%$, and the PQ score by $2.07\%$. On the BBBC006v1 database, it improves the mean APs by $1.52\%$, the AJI score by $1.62\%$, and the PQ score by $1.50\%$. The above experiments justify the effectiveness of CEM.
		
		\begin{table*}[!htp]
			\centering
			\caption{Ablation study on numbers of sampling points in CEM.}
			\label{tab:num_sp}
			\setlength{\tabcolsep}{3pt}
			\begin{tabular}{p{50pt}<{\centering} | p{55pt}<{\centering} p{50pt}<{\centering} p{50pt}<{\centering} p{50pt}<{\centering} p{50pt}<{\centering} p{50pt}<{\centering} p{50pt}<{\centering}}
				\hline
				Dataset & $N$ & $AP_{0.5}$ & $AP_{0.7}$ & $AP_{0.9}$ & $AP_{0.5:0.05:0.9}$ & AJI & PQ\\
				\hline
				\multirow{11}*{\makecell[c]{DSB2018\\val}} &0 & 0.8619 & 0.7047 & 0.2615 & 0.6454 & 0.8102 & 0.7749\\
				&1 & 0.8656 & 0.7180 & 0.3175 & 0.6691 & 0.8226 & 0.7901\\
				&2 & 0.8666 & 0.7159 & 0.3347 & 0.6702 & 0.8239 & 0.7910\\
				&3 & 0.8677 & 0.7210 & 0.3234 & 0.6739 & 0.8236 & 0.7931\\
				&4 & 0.8682 & 0.7255 & 0.3373 & 0.6762 & 0.8262 & 0.7938\\
				&5 & 0.8681 & 0.7244 & 0.3411 & 0.6756 & 0.8265 & 0.7944\\
				&6 & 0.8670 & 0.7303 & 0.3472 & \textbf{0.6799} & \textbf{0.8289} & 0.7956\\
				&7 & 0.8699 & 0.7233 & 0.3491 & 0.6788 & 0.8264 & 0.7952\\
				&8 & 0.8719 & 0.7211 & 0.3252 & 0.6711 & 0.8270 & 0.7942\\
				&9 & 0.8702 & 0.7157 & 0.3400 & 0.6754 & 0.8287 & 0.7950\\
				&10& 0.8679 & 0.7268 & 0.3441 & 0.6773 & 0.8265 & \textbf{0.7958}\\
				\hline
				\multirow{11}*{\makecell[c]{BBBC006v1\\val}} &0 & 0.9736 & 0.9449 & 0.8249 & 0.9295 & 0.9227 & 0.9326\\
				&1 & 0.9722 & 0.9456 & 0.8808 & 0.9399 & 0.9324 & 0.9409\\
				&2 & 0.9704 & 0.9460 & 0.8916 & 0.9410 & 0.9362 & 0.9451\\
				&3 & 0.9722 & 0.9475 & 0.8955 & 0.9439 & 0.9379 & 0.9468\\
				&4 & 0.9722 & 0.9473 & 0.8944 & 0.9425 & 0.9380 & 0.9472\\
				&5 & 0.9713 & 0.9485 & 0.8927 & 0.9431 & 0.9380 & 0.9463\\
				&6 & 0.9723 & 0.9498 & 0.8956 & \textbf{0.9447} & \textbf{0.9389} & \textbf{0.9476}\\
				&7 & 0.9726 & 0.9485 & 0.8964 & 0.9440 & 0.9381 & 0.9465\\
				&8 & 0.9708 & 0.9470 & 0.8920 & 0.9410 & 0.9374 & 0.9465\\
				&9 & 0.9721 & 0.9461 & 0.8971 & 0.9429 & 0.9381 & 0.9473\\
				&10& 0.9711 & 0.9484 & 0.8945 & 0.9436 & \textbf{0.9389} & 0.9471\\
				\hline
			\end{tabular}
		\end{table*}
		
		\subsection{Evaluation of CWM}
		
		The results of the ablation study on the CWM module are summarized in Table \ref{tab:srm}. In this table, `baseline' refers to the StarDist model \cite{stardist}, i.e., setting $N$ in CPP-Net to 0. In addition to CWM, another two weighting strategies are evaluated. `Equal weights' denotes the averaging weighting strategy for Eq. \eqref{eq:3.1}, while `Na\"{i}ve attention' represents learning fixed weights for the $N+1$ points in Eq. \eqref{eq:3.1}, using a trainable vector with $N+1$ elements.
		
		It is shown that CEM consistently outperforms the baseline model by large margins, regardless of the specific weighting strategy in Eq. \eqref{eq:3.1}. Moreover, compared to the other two weighting strategies, CWM achieves the best mean AP performance. CWM's advantage lies mainly in its APs under high IoU thresholds, which indicates that the instance segmentation accuracy is increased. This performance improvement can be ascribed to the superior flexibility of CWM. In short, unlike the two weighting strategies that adopt fixed weights, CWM can adaptively weigh each sampled point according to the quality of its features. The above experimental results justify the effectiveness of CWM.
		
		\begin{table*}[!htp]
			\centering
			\caption{Ablation study investigating different weighting strategies in CEM.}
			\label{tab:srm}
			\setlength{\tabcolsep}{3pt}
			\begin{tabular}{p{50pt}<{\centering} | p{55pt}<{\centering} p{50pt}<{\centering} p{50pt}<{\centering} p{50pt}<{\centering} p{50pt}<{\centering} p{50pt}<{\centering} p{50pt}<{\centering}}
				\hline
				Dataset & Method & $AP_{0.5}$ & $AP_{0.7}$ & $AP_{0.9}$ & $AP_{0.5:0.05:0.9}$ & AJI & PQ\\
				\hline
				\multirow{4}*{\makecell[c]{\\DSB2018\\val}}
				&\makecell[c]{baseline} & 0.8619 & 0.7047 & 0.2615 & 0.6454 & 0.8102 & 0.7749\\
				& \makecell[c]{equal weights} & 0.8636 & 0.7173 & 0.3223 & 0.6691 & 0.8209 & 0.7910\\
				& \makecell[c]{na\"{i}ve attention} & 0.8596 & 0.7175 & 0.3277 & 0.6669 & 0.8211 & 0.7886\\
				& \makecell[c]{CWM} & 0.8670 & 0.7303 & 0.3472 & \textbf{0.6799} & \textbf{0.8289} & \textbf{0.7956}\\
				\hline
				\multirow{4}*{\makecell[c]{\\BBBC006v1\\val}}
				&\makecell[c]{baseline} & 0.9736 & 0.9449 & 0.8249 & 0.9295 & 0.9227 & 0.9326\\
				& \makecell[c]{equal weights} & 0.9753 & 0.9502 & 0.8851 & 0.9434 & 0.9356 & 0.9468\\
				& \makecell[c]{na\"{i}ve attention} & 0.9727 & 0.9493 & 0.8947 & 0.9443 & 0.9378 & 0.9469\\
				& \makecell[c]{CWM} & 0.9723 & 0.9498 & 0.8956 & \textbf{0.9447} & \textbf{0.9389} & \textbf{0.9476}\\
				\hline
			\end{tabular}
		\end{table*}
		
		\subsection{Evaluation of the SAP Loss}
		
		In this experiment, we justify the effectiveness of the SAP loss. Utilizing the SAP loss requires training an encoder-decoder model that transforms the instance representations in CPP-Net to other types of representations (as described in Section \ref{sec:saploss}). Accordingly, we evaluate the following three types of representation strategies for the SAP loss. The first strategy is boundary-based, in that it predicts both semantic segmentation masks and instance boundaries \cite{dcan, bes-net, cia-net}; the second strategy is bounding box-based, in that it regresses both the coordinates of nucleus centroids and bounding box positions for each pixel inside one instance \cite{spa-net}. The third strategy predicts both the above mentioned representations. For simplicity, these three strategies are denoted as `seg \& bnd', `bbox', and `both' in Table \ref{tab:iap_loss}.
		
		In Table \ref{tab:iap_loss}, we first show the performance of CPP-Net without using the SAP loss. On both datasets, the SAP loss promotes performance in terms of mean AP. Specifically, the SAP loss improves the mean AP by $0.90\%$, the AJI score by $0.31\%$, and the PQ score by $0.59\%$ on DSB2018. It improves the mean AP by $0.50\%$, the AJI score by $0.35\%$, and the PQ score by $0.34\%$ on BBBC006v1. Furthermore, it is also clear that the improvement is mainly from APs under high IoU thresholds: for example, $1.96\%$ improvements on $AP_{0.9}$ on DSB2018 and $1.32\%$ improvements on $AP_{0.9}$ on BBBC006v1. For APs with lower IoU values, SAP loss does not introduce significant performance promotion. From this phenomenon, we can conclude that the SAP loss primarily penalizes the prediction error in nucleus shape, rather than the localization or detection errors.
		
		We also train the CPP-Net with another variant of the SAP loss, in which the encoder-decoder model is trained to reconstruct its input representations, i.e., the ground-truth centroid probability and pixel-to-boundary distance maps. The results of CPP-Net trained with this variant are denoted as `recons.' in Table \ref{tab:iap_loss}. The results show that the proposed SAP loss achieves better performance than this variant. The advantage achieved by our proposed SAP loss can be attributed to the transformation between different representation strategies. Through the use of this transformation task, the encoder-decoder model is forced to extract essential information related to the nucleus shape. By contrast, the `recons.' variant is likely to only memorize the input information. Accordingly, our proposed SAP loss achieves better overall performance than all other three variants. In the following, we adopt our proposed SAP loss to train CPP-Net.
		
		\begin{table*}[!htp]
			\caption{Ablation study investigating the Shape-Aware Perceptual (SAP) loss.}
			\label{tab:iap_loss}
			\centering
			\setlength{\tabcolsep}{3pt}
			\begin{tabular}{p{50pt}<{\centering} | p{55pt} p{50pt}<{\centering} p{50pt}<{\centering} p{50pt}<{\centering} p{50pt}<{\centering} p{50pt}<{\centering} p{50pt}<{\centering}}
				\hline
				Dataset & \makecell[c]{SAP loss} & $AP_{0.5}$ & $AP_{0.7}$ & $AP_{0.9}$ & $AP_{0.5:0.05:0.9}$ & $AJI$ & $PQ$ \\
				\hline
				\multirow{5}*{\makecell[c]{\\DSB2018\\val}} &\makecell[c]{-} & 0.8670 & 0.7303 & 0.3472 & 0.6799 & 0.8289 & 0.7956\\
				&\makecell[c]{seg \& bnd} & 0.8712 & 0.7358 & 0.3517 & 0.6840 & 0.8265 & 0.7990\\
				&\makecell[c]{bbox} & 0.8719 & 0.7259 & 0.3320 & 0.6806 & 0.8303 & 0.7976\\
				&\makecell[c]{both} & 0.8748 & 0.7347 & 0.3668 & \textbf{0.6889} & \textbf{0.8320} & \textbf{0.8015}\\
				&\makecell[c]{recons.} & 0.8622 & 0.7405 & 0.3571 & 0.6873 & 0.8278 & 0.7972\\
				\hline
				\multirow{5}*{\makecell[c]{\\BBBC006v1\\val}} &\makecell[c]{-} & 0.9723 & 0.9498 & 0.8956 & 0.9447 & 0.9389 & 0.9476\\
				&\makecell[c]{seg \& bnd} & 0.9757 & 0.9531 & 0.9038 & 0.9483 & 0.9395 & 0.9488\\
				&\makecell[c]{bbox} & 0.9731 & 0.9508 & 0.8978 & 0.9453 & 0.9381 & 0.9466\\
				&\makecell[c]{both} & 0.9761 & 0.9530 & 0.9088 & \textbf{0.9497} & \textbf{0.9424} & \textbf{0.9510}\\
				&\makecell[c]{recons.} & 0.9740 & 0.9479 & 0.8908 & 0.9438 & 0.9359 & 0.9461\\
				\hline
			\end{tabular}
		\end{table*}
		
		\begin{table*}[!htp]
			\caption{Ablation study investigating the Fine-grained Post-Processing (FPP) pipeline.}
			\label{tab:spp}
			\centering
			\setlength{\tabcolsep}{3pt}
			\begin{tabular}{p{50pt}<{\centering} | p{55pt} p{50pt}<{\centering} p{50pt}<{\centering} p{50pt}<{\centering} p{50pt}<{\centering} p{50pt}<{\centering} p{50pt}<{\centering}}
				\hline
				Dataset & \makecell[c]{Post-processing} & $AP_{0.5}$ & $AP_{0.7}$ & $AP_{0.9}$ & $AP_{0.5:0.05:0.9}$ & $AJI$ & $PQ$ \\
				\hline
				\multirow{2}*{\makecell[c]{DSB2018\\val}} &\makecell[c]{original} & 0.8748 & 0.7347 & 0.3668 & 0.6889 & 0.8320 & 0.8015\\
				&\makecell[c]{FPP} & 0.8724 & 0.7486 & 0.4169 & \textbf{0.7037} & \textbf{0.8336} & \textbf{0.8118}\\
				\hline
				\multirow{2}*{\makecell[c]{BBBC006v1\\val}} &\makecell[c]{original} & 0.9750 & 0.9514 & 0.9048 & 0.9481 & 0.9401 & 0.9490\\
				&\makecell[c]{FPP} & 0.9814 & 0.9568 & 0.9258 & \textbf{0.9561} & \textbf{0.9628} & \textbf{0.9724}\\
				\hline
			\end{tabular}
		\end{table*}

		\subsection{Evaluation of FPP}
		The results of adopting the original post-processing method in StarDist \cite{stardist} and FPP are presented in Table \ref{tab:spp}. On DSB2018, FPP improves the mean AP by $1.48\%$, the AJI by $0.16\%$, and the PQ by $1.03\%$. On BBBC006v1, it improves the mean AP by $0.80\%$, the AJI by $2.27\%$, and the PQ by $2.34\%$. Moreover, according to Table \ref{tab:spp}, FPP improves $AP_{0.9}$ by $5.01\%$ on DSB2018 and $2.1\%$ on BBBC006v1. This is because FPP effectively refines the boundaries of nucleus instances, which considerably improves the segmentation quality.
		
		\subsection{Comparisons with State-of-the-Art Methods}
		
		\begin{table*}[htbp]
			\centering
			\caption{Comparisons with SOTA methods on DSB2018 and BBBC006v1. * denotes methods evaluated by ourselves.}
			\label{tab:cmp_0}
			\setlength{\tabcolsep}{3pt}
			\begin{tabular}{p{50pt}<{\centering} | p{85pt} p{35pt}<{\centering} p{35pt}<{\centering} p{35pt}<{\centering} p{50pt}<{\centering} p{50pt}<{\centering} p{50pt}<{\centering}}
				\hline
				Dataset & \makecell[c]{Methods} & $AP_{0.5}$ & $AP_{0.7}$ & $AP_{0.9}$ & $AP_{0.5:0.05:0.9}$ & $AJI$ & $PQ$ \\
				\hline
				\multirow{9}*{\makecell[c]{DSB2018\\test}}
				&\makecell[c]{ Ellipse Fitting* \cite{icip18}}    & 0.5903 & 0.3485 & 0.0427 & 0.3350$\pm$0.1491 & 0.5626$\pm$0.1447 & 0.5454$\pm$0.1389 \\
				&\makecell[c]{ Watershed* \cite{cellsegreview}}   & 0.6977 & 0.5039 & 0.2154 & 0.4838$\pm$0.2531 & 0.6753$\pm$0.1842 & 0.6497$\pm$0.1901\\
				&\makecell[c]{Mask R-CNN \cite{stardist}}    & 0.8323 & 0.6838 & 0.1891 & 0.6058 & - &  - \\
				&\makecell[c]{StarDist \cite{stardist}}       & 0.8641 & 0.6850 & 0.1191 & 0.5983 & - &  - \\
				&\makecell[c]{PatchPerPix \cite{patchperpix}} & 0.8680 & 0.7550 & 0.3790 & 0.7046 & - & - \\
				&\makecell[c]{KeypointGraph* \cite{keygraph}} & 0.8244 & 0.7083 & 0.2989 & 0.6561$\pm$0.1782 & 0.8014$\pm$0.1009 & 0.7812$\pm$0.0996\\
				&\makecell[c]{HoVer-Net* \cite{hovernet}}    & 0.7838 & 0.7165 & 0.3978 & 0.6611$\pm$0.2296 & 0.7752$\pm$0.1583 & 0.7640$\pm$0.1607\\
				&\makecell[c]{StarDist* \cite{stardist}}      & 0.8731 & 0.7368 & 0.2566 & 0.6657$\pm$0.1849 & 0.8088$\pm$0.0955 & 0.7842$\pm$0.1006\\
				&\makecell[c]{CPP-Net*}                       & 0.8821 & 0.7887 & 0.3856 & \textbf{0.7228}$\pm$0.1872 & \textbf{0.8278}$\pm$0.0972 & \textbf{0.8158}$\pm$0.1012\\
				\hline
				\multirow{7}*{\makecell[c]{\\BBBC006v1\\test}}
				&\makecell[c]{ Ellipse Fitting* \cite{icip18}}    & 0.7456 & 0.5353 & 0.0174 & 0.4530$\pm$0.0810 & 0.6911$\pm$0.0518 & 0.6592$\pm$0.0638 \\
				&\makecell[c]{ Watershed* \cite{cellsegreview}}    & 0.8071 & 0.7582 & 0.1182 & 0.6524$\pm$0.0857 & 0.7818$\pm$0.0529 & 0.7624$\pm$0.0582\\
				&\makecell[c]{InstanceEmbedding* \cite{instanceembedding}} & 0.8327 & 0.7862 & 0.0258 & 0.6712$\pm$0.0680 & 0.7504$\pm$0.0394 & 0.7728$\pm$0.0366\\
				&\makecell[c]{KeypointGraph* \cite{keygraph}} & 0.9365 & 0.8927 & 0.0879 & 0.7594$\pm$0.0536 & 0.8199$\pm$0.0334 & 0.8294$\pm$0.0266\\
				&\makecell[c]{HoVer-Net* \cite{hovernet}} & 0.9268 & 0.8962 & 0.8676 & 0.8969$\pm$0.0598 & 0.9215$\pm$0.0412 & 0.9261$\pm$0.0309\\
				&\makecell[c]{StarDist* \cite{stardist}} & 0.9757 & 0.9503 & 0.8189 & 0.9304$\pm$0.0486 & 0.9235$\pm$0.0264 & 0.9321$\pm$0.0238\\
				&\makecell[c]{CPP-Net*} & 0.9811 & 0.9584 & 0.9231 & \textbf{0.9548}$\pm$0.0407 & \textbf{0.9624}$\pm$0.0274 & \textbf{0.9709}$\pm$0.0229\\
				\hline
			\end{tabular}
		\end{table*}
		
		\begin{figure*}
			\centering
			\includegraphics[width=1.0\linewidth]{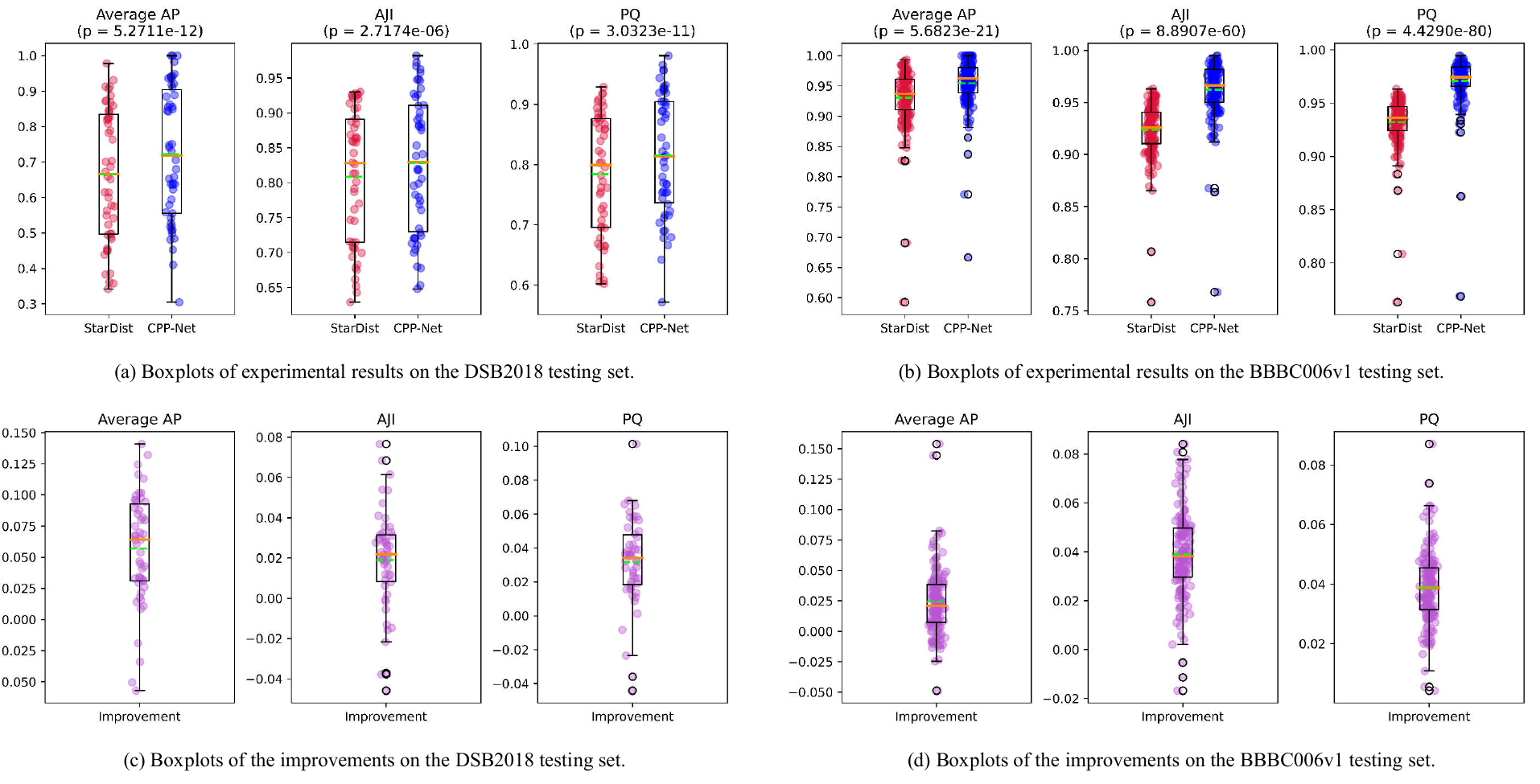}
			\caption{Boxplots of the mean AP, AJI, and PQ scores on two datasets. (a) and (b) present the scores of StarDist and CPP-Net, respectively. (c) and (d) illustrate the improvements achieved by CPP-Net relative to StarDist. The orange and green lines in each box indicate the median and average values, respectively. Each testing sample is represented as one point in the figure. T-test is used to compare the results of StarDist and CPP-Net. The p-value is presented for each figure of (a) and (b). Best viewed in color.}
			\label{fig:boxplot}
		\end{figure*}
		
		\begin{table}[htbp]
			\centering
			\caption{Average inference time on the DSB2018 database.}
			\label{tab:inf_time}
			\setlength{\tabcolsep}{3pt}
			\begin{tabular}{p{100pt}<{\centering} p{120pt}<{\centering}}
				\hline
				\makecell[c]{Methods} & \makecell[c]{Average Inference Time \\ (second per image)}\\
				\hline
				\makecell[c]{KeypointGraph \cite{keygraph}} & \makecell[c]{0.8556}  \\
				\makecell[c]{HoVer-Net \cite{hovernet}} & \makecell[c]{1.5556} \\
				\makecell[c]{PatchPerPix \cite{patchperpix}} & \makecell[c]{5.8767} \\
				\makecell[c]{StarDist \cite{stardist}} &  \makecell[c]{0.2327} \\
				\makecell[c]{CPP-Net w/o FPP} & \makecell[c]{0.2519}\\
				\makecell[c]{CPP-Net w/ FPP} & \makecell[c]{0.2609}\\
				\hline
			\end{tabular}
		\end{table}
		
		\begin{table*}[htbp]
			\centering
			\caption{Comparisons with SOTA methods on the PanNuke database. * denotes methods evaluated by ourselves.}
			\label{tab:cmp_1}
			\setlength{\tabcolsep}{3pt}
			\begin{tabular}{p{75pt} p{25pt}<{\centering} p{25pt}<{\centering} p{25pt}<{\centering} p{25pt}<{\centering} p{25pt}<{\centering} p{25pt}<{\centering} p{25pt}<{\centering} p{25pt}<{\centering} p{25pt}<{\centering} p{25pt}<{\centering} p{25pt}<{\centering} p{25pt}<{\centering} p{25pt}<{\centering} p{25pt}<{\centering}}
				\hline
				Tissue & \multicolumn{2}{c}{Mask R-CNN \cite{pannuke_extend}} & \multicolumn{2}{c}{Micro-Net \cite{pannuke_extend}} & \multicolumn{2}{c}{HoVer-Net \cite{pannuke_extend}} & \multicolumn{2}{c}{StarDist* \cite{stardist}} & \multicolumn{2}{c}{CPP-Net*} & \multicolumn{2}{c}{\makecell[c]{StarDist* \cite{stardist} \\ with ResNet50}} & \multicolumn{2}{c}{\makecell[c]{CPP-Net* \\ with ResNet50}}\\
				\hline
				\quad & mPQ & bPQ & mPQ & bPQ & mPQ & bPQ & mPQ & bPQ & mPQ & bPQ & mPQ & bPQ & mPQ & bPQ\\
				Adrenal Gland 			& 0.3470 & 0.5546 & 0.4153 & 0.6440 & 0.4812 & 0.6962 & 0.4855 & 0.6764 & 0.4856 & 0.6974 & 0.4868 & 0.6972 & 0.4944 & 0.7066\\
				Bile Duct 				& 0.3536 & 0.5567 & 0.4124 & 0.6232 & 0.4714 & 0.6696 & 0.4492 & 0.6417 & 0.4564 & 0.6619 & 0.4651 & 0.6690 & 0.4670 & 0.6768\\
				Bladder 				& 0.5065 & 0.6049 & 0.5357 & 0.6488 & 0.5792 & 0.7031 & 0.5718 & 0.6798 & 0.5903 & 0.6866 & 0.5793 & 0.6986 & 0.5936 & 0.7053\\
				Breast 					& 0.3882 & 0.5574 & 0.4407 & 0.6029 & 0.4902 & 0.6470 & 0.4946 & 0.6507 & 0.5080 & 0.6649 & 0.5064 & 0.6666 & 0.5090 & 0.6747\\
				Cervix 					& 0.3402 & 0.5483 & 0.3795 & 0.6101 & 0.4438 & 0.6652 & 0.4544 & 0.6659 & 0.4638 & 0.6780 & 0.4628 & 0.6690 & 0.4792 & 0.6912\\
				Colon 					& 0.3122 & 0.4603 & 0.3414 & 0.4972 & 0.4095 & 0.5575 & 0.4009 & 0.5534 & 0.4163 & 0.5714 & 0.4205 & 0.5779 & 0.4315 & 0.5911\\
				Esophagus 				& 0.4311 & 0.5691 & 0.4668 & 0.6011 & 0.5085 & 0.6427 & 0.5206 & 0.6465 & 0.5333 & 0.6634 & 0.5331 & 0.6655 & 0.5449 & 0.6797\\
				Head \& Neck 			& 0.3946 & 0.5457 & 0.3668 & 0.5242 & 0.4530 & 0.6331 & 0.4613 & 0.6331 & 0.4646 & 0.6337 & 0.4768 & 0.6433 & 0.4706 & 0.6523\\
				Kidney 					& 0.3553 & 0.5092 & 0.4165 & 0.6321 & 0.4424 & 0.6836 & 0.4902 & 0.6802 & 0.4835 & 0.6972 & 0.4880 & 0.6998 & 0.5194 & 0.7067\\
				Liver 					& 0.4103 & 0.6085 & 0.4365 & 0.6666 & 0.4974 & 0.7248 & 0.4891 & 0.7007 & 0.5000 & 0.7212 & 0.5145 & 0.7231 & 0.5143 & 0.7312\\
				Lung 					& 0.3182 & 0.5134 & 0.3370 & 0.5588 & 0.4004 & 0.6302 & 0.4032 & 0.6165 & 0.4110 & 0.6288 & 0.4128 & 0.6362 & 0.4256 & 0.6386\\
				Ovarian 				& 0.4337 & 0.5784 & 0.4387 & 0.6013 & 0.4863 & 0.6309 & 0.5170 & 0.6499 & 0.5200 & 0.6729 & 0.5205 & 0.6668 & 0.5313 & 0.6830\\
				Pancreatic 				& 0.3624 & 0.5460 & 0.4041 & 0.6074 & 0.4600 & 0.6491 & 0.4410 & 0.6331 & 0.4815 & 0.6597 & 0.4585 & 0.6601 & 0.4706 & 0.6789\\
				Prostate 				& 0.3959 & 0.5789 & 0.4341 & 0.6049 & 0.5101 & 0.6615 & 0.4998 & 0.6473 & 0.5176 & 0.6735 & 0.5067 & 0.6748 & 0.5305 & 0.6927\\
				Skin 					& 0.2665 & 0.5021 & 0.3223 & 0.5817 & 0.3429 & 0.6234 & 0.3537 & 0.6063 & 0.3420 & 0.6041 & 0.3610 & 0.6289 & 0.3574 & 0.6209\\
				Stomach 				& 0.3684 & 0.5976 & 0.3872 & 0.6293 & 0.4726 & 0.6886 & 0.4191 & 0.6636 & 0.4420 & 0.6987 & 0.4477 & 0.6944 & 0.4582 & 0.7067\\
				Testis 					& 0.3512 & 0.5420 & 0.4088 & 0.6300 & 0.4754 & 0.6890 & 0.4767 & 0.6661 & 0.4943 & 0.6860 & 0.4942 & 0.6869 & 0.4931 & 0.7026\\
				Thyroid 				& 0.3037 & 0.5712 & 0.3712 & 0.6555 & 0.4315 & 0.6983 & 0.4166 & 0.6807 & 0.4509 & 0.7127 & 0.4300 & 0.6962 & 0.4392 & 0.7155\\
				Uterus					& 0.3683 & 0.5589 & 0.3965 & 0.5821 & 0.4393 & 0.6393 & 0.4428 & 0.6305 & 0.4604 & 0.6473 & 0.4480 & 0.6599 & 0.4794 & 0.6615\\
				\hline
				Average across tissues		& 0.3688 & 0.5528 & 0.4059 & 0.6053 & 0.4629 & 0.6596 & 0.4625 & 0.6485 & 0.4748 & 0.6663 & 0.4744 & 0.6692 & \textbf{0.4847} & \textbf{0.6798} \\
				STD across splits		& 0.0047 & 0.0076 & 0.0082 & 0.0050 & 0.0076 & 0.0036 & 0.0078 & 0.0054 & 0.0068 & 0.0051 & 0.0037 & 0.0014 & 0.0059 & 0.0015 \\
				\hline
			\end{tabular}
		\end{table*}
		
		\begin{table}[htbp]
			\centering
			
			\caption{P values of the comparison between StarDist and CPP-Net on the PanNuke database. T-test is used.}
			\label{tab:pannuke_pvalue}
			\setlength{\tabcolsep}{3pt}
			\begin{tabular}{p{50pt}<{\centering} p{70pt}<{\centering} p{70pt}<{\centering}}
				\hline
				\makecell[c]{Backbone} & \makecell[c]{P value of mPQ} & \makecell[c]{P value of bPQ}\\
				\hline
				\makecell[c]{UNet} & \makecell[c]{5.2833e-5} & \makecell[c]{5.5250e-14}\\
				\makecell[c]{ResNet50} & \makecell[c]{0.0269} & \makecell[c]{1.2558e-6}\\
				\hline
			\end{tabular}
		\end{table}

		\subsubsection{Comparisons on the DSB2018 database}
		
		We compare the performance of CPP-Net with both traditional methods and deep-learning based methods. More specifically, we evaluate the method in \cite{icip18} that fits each cell with one ellipse; in addition, we also evaluate the watershed-based method following the pipeline in \cite{cellsegreview}. These two traditional methods are denoted as `Ellipse Fitting' and `Watershed' respectively in Table \ref{tab:cmp_0}. Furthermore, we compare the performance of CPP-Net with Mask-RCNN\cite{stardist,mask-rcnn}, KeypointGraph\cite{keygraph}, HoVer-Net\cite{hovernet}, PatchPerPix\cite{patchperpix}, and StarDist\cite{stardist}. The results of this comparison are tabulated in Table \ref{tab:cmp_0}. It is notable here that some above-mentioned methods were evaluated using different training and testing data split protocols in their respective papers. In the interests of fair comparison, we evaluate the performance of HoVer-Net \cite{hovernet} and KeypointGraph \cite{keygraph} by ourselves using codes released by the authors, under the same evaluation protocol as \cite{stardist,patchperpix}. We also re-implement the StarDist approach on DSB2018 and replace its BN layers with GN layers. Accordingly, we achieve better performances than the results reported in \cite{stardist}.
		
		As shown in Table \ref{tab:cmp_0}, StarDist and PatchPerPix are two powerful approaches and have their own respective advantages. Specifically, StarDist achieves higher $AP_{0.5}$ than PatchPerPix, but much lower APs under high IoU thresholds. We conjecture the StarDist may be affected by prediction accuracy regarding the shape of nucleus boundaries. This is because StarDist adopts the features of centroid pixels only for shape prediction; however, the centroid pixel alone lacks contextual information. In comparison, CPP-Net consistently achieves better performance than StarDist; in particular, it improves the performance at high IoU thresholds. Finally, CPP-Net achieves the best mean AP performance among all methods. The above comparison experiments justify the effectiveness of CPP-Net.
		
		The boxplots in Fig. \ref{fig:boxplot}(a) and \ref{fig:boxplot}(c) present a comparison between StarDist and CPP-Net. As shown in Fig. \ref{fig:boxplot}(a), the quartiles of CPP-Net are higher than those of StarDist. Moreover, CPP-Net has a higher upper whisker than StarDist and improves the segmentation results on most testing samples. Moreover, t-test is used to compare the results of StarDist and CPP-Net. The p values are presented in Fig. \ref{fig:boxplot}. They are all smaller than 0.05, which means that the improvement achieved by CPP-Net is statistically significant.
		
		We further summarize the inference time of different models in Table \ref{tab:inf_time}. Here, inference time includes the network prediction time and the associated post-processing time. We compare the inference time under the same hardware conditions: one NVIDIA TITAN Xp GPU, Intel(R) Core(TM) i7-6850K CPU @3.60GHz, and 128GB RAM. As shown in Table \ref{tab:inf_time}, StarDist \cite{stardist} is the fastest among all compared approaches.
		
		With the same post-processing pipeline, CPP-Net increases the time costs by only 0.0282 seconds per image compared with StarDist. While the FPP method introduces an increasing time cost, the overall time cost of CPP-Net is still much smaller than the majority of existing methods \cite{hovernet,keygraph,patchperpix}. Therefore, compared with most approaches presented in Table \ref{tab:inf_time}, CPP-Net is highly efficient.
		
		\begin{table*}[htbp]
			\centering
			\caption{Cross-dataset Evaluation. * denotes methods evaluated by ourselves.}
			\label{tab:cross_eval}
			\setlength{\tabcolsep}{3pt}
			\begin{tabular}{p{85pt}<{\centering} | p{50pt}<{\centering} p{35pt}<{\centering} p{35pt}<{\centering} p{35pt}<{\centering} p{50pt}<{\centering} p{50pt}<{\centering} p{50pt}<{\centering}}
				\hline
				\makecell[c]{Tasks} & \makecell[c]{Methods} & $AP_{0.5}$ & $AP_{0.7}$ & $AP_{0.9}$ & $AP_{0.5:0.05:0.9}$ & $AJI$ & $PQ$ \\
				\hline
				\multirow{2}*{\makecell[c]{DSB2018$\rightarrow$BBBC039}} &\makecell[c]{StarDist* \cite{stardist}} & 0.9090 & 0.8492 & 0.4168 & 0.7825$\pm$0.1017 & 0.8644$\pm$0.0707 & 0.8459$\pm$0.0622\\
				&\makecell[c]{CPP-Net*} & 0.9081 & 0.8783 & 0.6280 & \textbf{0.8436}$\pm$0.0920 & \textbf{0.8902}$\pm$0.0686 & \textbf{0.8808}$\pm$0.0577\\
				\hline
				\multirow{2}*{\makecell[c]{DSB2018$\rightarrow$BBBC006v1}} &\makecell[c]{StarDist* \cite{stardist}} & 0.7629 & 0.6537 & 0.0685 & 0.5483$\pm$0.0893 & 0.7362$\pm$0.0521 & 0.7101$\pm$0.0627\\
				&\makecell[c]{CPP-Net*} & 0.8077 & 0.7118 & 0.2009 & \textbf{0.6375}$\pm$0.0904 & \textbf{0.7756}$\pm$0.0525 & \textbf{0.7660}$\pm$0.0577\\
				\hline
				\multirow{2}*{\makecell[c]{BBBC006v1$\rightarrow$BBBC039}} &\makecell[c]{StarDist* \cite{stardist}} & 0.7907 & 0.7248 & 0.2038 & 0.6436$\pm$0.1110 & 0.7633$\pm$0.0863 & 0.7594$\pm$0.0914\\
				&\makecell[c]{CPP-Net*} & 0.7813 & 0.7249 & 0.2898 & \textbf{0.6632}$\pm$0.1084 & \textbf{0.7699}$\pm$0.0873 & \textbf{0.7701}$\pm$0.0882\\
				\hline
				\multirow{2}*{\makecell[c]{BBBC006v1$\rightarrow$DSB2018}} &\makecell[c]{StarDist* \cite{stardist}} & 0.3668 & 0.2628 & 0.0647 & 0.2407$\pm$0.2134 & 0.4342$\pm$0.2335 & 0.3742$\pm$0.2644\\
				&\makecell[c]{CPP-Net*} & 0.3754 & 0.2792 & 0.1047 & \textbf{0.2624}$\pm$0.2254 & \textbf{0.4419}$\pm$0.2362 & \textbf{0.3950}$\pm$0.2598\\
				\hline
			\end{tabular}
			
		\end{table*}

		\subsubsection{Comparisons on the BBBC006v1 database}
		
		To facilitate fair comparison, we train StarDist \cite{stardist}, HoVer-Net \cite{hovernet}, KeypointGraph \cite{keygraph}, and InstanceEmbedding \cite{instanceembedding} using the same data split protocol as ours. Experimental results are summarized in Table \ref{tab:cmp_0}. As the table shows, similar to the results on DSB2018, the StarDist model achieves a promising $AP_{0.5}$ score but an unsatisfactory $AP_{0.9}$ score. By contrast, the proposed CPP-Net promotes the nucleus segmentation performance and maintains its advantages in terms of nucleus detection. It also continues to outperform all the other state-of-the-art methods. We further draw boxplots from the experimental results on BBBC006v1 in Fig. \ref{fig:boxplot}(b) and \ref{fig:boxplot}(d), which clearly illustrate the advantages of CPP-Net. Experimental results on this database justify the effectiveness of CPP-Net.
		
		\subsubsection{Comparisons on the PanNuke database}
		
		We provide the performance of StarDist and CPP-Net with two different backbones. The first backbone adopts the same encoder as that used in the DSB2018 database, while the second employs ResNet-50 as the encoder. Their performance is compared with that of Mask-RCNN \cite{mask-rcnn}, Micro-Net \cite{micro-net}, and HoVer-Net \cite{hovernet} in Table \ref{tab:cmp_1}. We further adopt the same evaluation metrics as those in \cite{pannuke_extend}, where we also copied the results of Mask-RCNN \cite{mask-rcnn}, Micro-Net \cite{micro-net}, and HoVer-Net \cite{hovernet}. In Table \ref{tab:cmp_1}, both bPQ and mPQ are computed for each of the 19 tissues.
		
		As the experimental results in Table \ref{tab:cmp_1} demonstrate, CPP-Net consistently outperforms StarDist using each of the two backbones. Moreover, when CPP-Net is equipped with the same ResNet-50 backbone as HoVer-Net, it achieves better average performance than all other methods: for example, it outperforms StarDist by $1.03\%$ and $1.06\%$ in mPQ and bPQ, respectively. Moreover, t-test is used to compare the results of StarDist and CPP-Net, and the p values are listed in Table \ref{tab:pannuke_pvalue}. The p values of the comparison between StarDist and CPP-Net are smaller than 0.05. Results of the above comparisons are consistent with those on the first two databases, which further justifies the effectiveness of CPP-Net.

		\subsubsection{Cross-dataset evaluation}
		To justify the generalization ability of CPP-Net, we further conduct cross-dataset evaluations. More specifically, we test the performance of CPP-Net and StarDist on the DSB2018$\rightarrow$BBBC039, DSB2018$\rightarrow$BBBC006v1, BBBC006v1$\rightarrow$BBBC039, and BBBC006v1$\rightarrow$DSB2018 tasks. Each dataset on the left of the arrow stands for the training set, while the dataset on the right is used for testing. Experimental results of cross-dataset evaluation are presented in Table \ref{tab:cross_eval}.
		
		The DSB2018 database contains cells of various types; therefore, models trained on this database have better generalization ability. In comparison, BBBC006v1 includes only U2OS cells, and models trained on this dataset have limited generalization ability. For the four tasks, CPP-Net consistently achieves better performance than StarDist for mean AP, AJI, and PQ metrics. For example, on the DSB2018$\rightarrow$BBBC039 task, CPP-Net improves the mean AP by $6.11\%$, the AJI by $2.58\%$, and the PQ by $3.49\%$. These experimental results demonstrate the robustness and the generalization ability of CPP-Net.
		
		\begin{figure}[htbp]
			\centering
			\includegraphics[width=1.0\linewidth]{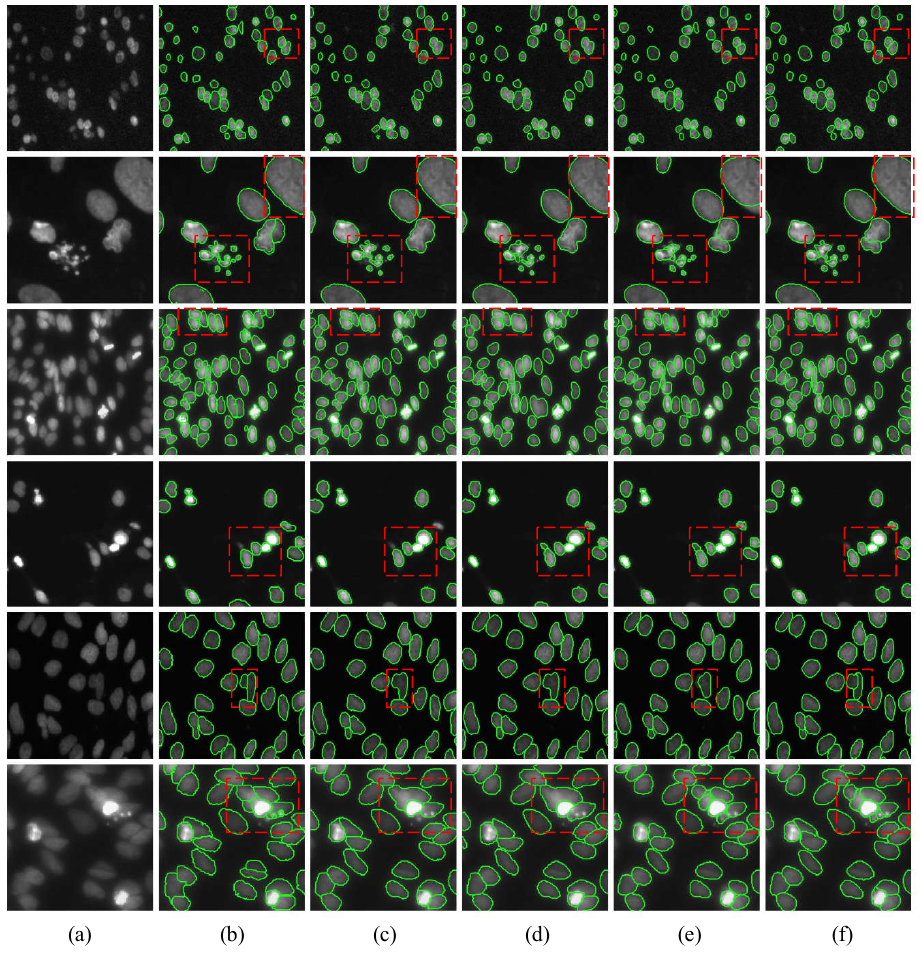}
			\caption{Qualitative comparisons between SOTA methods on the DSB2018 dataset. The six columns from left to right are the original images (a), the ground truth segmentation results (b), and predictions by HoVer-Net \cite{hovernet}, PatchPerPix \cite{patchperpix}, StarDist \cite{stardist} and CPP-Net (c-f). Best viewed with zoom-in.}
			\label{fig:vis_dsb}
		\end{figure}
		\begin{figure}[htbp]
			\centering
			\includegraphics[width=1.0\linewidth]{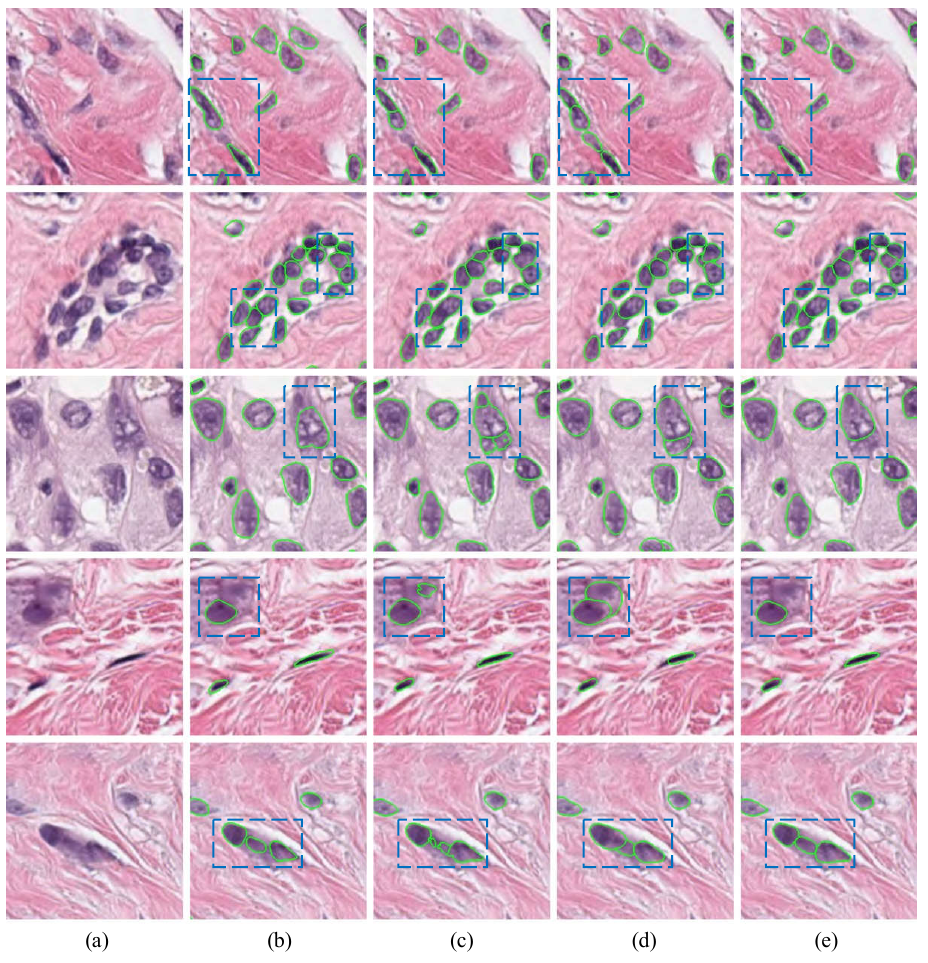}
			\caption{Qualitative comparisons between SOTA methods on the PanNuke dataset. The five columns from left to right are the original images (a), the ground truth segmentation results (b), and predictions by HoVer-Net \cite{hovernet}, StarDist \cite{stardist} and CPP-Net (c-f). Best viewed with zoom-in.}
			\label{fig:vis_pannuke}
		\end{figure}

		\subsection{Qualitative Comparisons}
		
		In this experiment, we conduct qualitative comparisons on the DSB2018 and PanNuke datasets. The results of different methods on the two datasets are presented in Fig. \ref{fig:vis_dsb} and Fig. \ref{fig:vis_pannuke}, respectively. In Fig. \ref{fig:vis_dsb}, we compare the results achieved by CPP-Net with HoVer-Net, PatchPerPix and StarDist on six examples in the DSB2018 test dataset. From Fig. \ref{fig:vis_dsb}, we have the following observations. First, compared with StarDist, CPP-Net performs better in terms of segmentation accuracy, e.g., the highlighted nuclei in the second rows. Second, compared with HoVer-Net, PatchPerPix and StarDist, CPP-Net is more powerful in separating touching nucleus instances, e.g., the highlighted nuclei in the first, fifth, and sixth rows. In Fig. \ref{fig:vis_pannuke}, CPP-Net is compared with HoVer-Net and StarDist on five examples in the PanNuke dataset. Similar observations can be found in Fig. \ref{fig:vis_pannuke}. For example, CPP-Net correctly distinguishes those touching nuclei in the second and fifth rows.
		
		\subsection{Limitation and Future Work}
		
		Similar to StarDist \cite{stardist}, CPP-Net is also built on the assumption that each nucleus instance can be represented by a convex polygon. While this is true for the vast majority of nuclei, it may not hold for nuclei with irregular shapes. One possible solution is to replace polygon with spline \cite{splinedist} and rebuild the CPP-Net on the spline model.
		
		In the future, we will apply the proposed method to related medical image analysis tasks. For example, since CPP-Net effectively aggregates context information, it can be applied to the nucleus classification task that relies on the global feature of one instance. Furthermore, since CEM and the SAP loss improve the shape quality of predicted foreground objects, we will try to apply them to brain tumor \cite{brats,ets_brats,cm_brats} and pancreas segmentation \cite{pancreas} tasks that requires precise segmentation quality.
		
		\section{conclusion}
		\label{sec:conclusion}
		
		In this paper, we improve the performance of StarDist from three aspects. First, we propose a Context Enhancement Module that enables us to explore more contextual information and accordingly predict the centroid-to-boundary distances more robustly, especially for large-sized  nuclei. We further propose a Confidence-based Weighting Module that adaptively fuses the predictions of the sampled points in the CEM module. Second, we propose a Shape-Aware Perceptual loss, which constrains the high-level shape information contained in the centroid probability and pixel-to-boundary distance maps. Third, we introduce Fine-grained Post-Processing method to refine the boundaries of nucleus instances. We conduct extensive ablation studies to justify the effectiveness of each proposed component. Finally, our proposed CPP-Net model outperforms the StarDist model and achieves state-of-the-art performance on the popular datasets for nucleus segmentation.

\end{document}